\pdfoutput=1

\documentclass[11pt]{article}

\usepackage[preprint]{acl}

\usepackage{times}
\usepackage{latexsym}
\usepackage{arydshln}
\usepackage{booktabs}

\usepackage[T1]{fontenc}

\usepackage[utf8]{inputenc}

\usepackage{microtype}

\usepackage{inconsolata}

\usepackage{graphicx}

\title{Trustworthy Hate Speech Detection Through Visual Augmentation}

\author{
 \textbf{Ziyuan Yang\textsuperscript{1,2,3}},
 \textbf{Ming Yan\textsuperscript{3,*}},
 \textbf{Yingyu Chen\textsuperscript{1,2}},
 \textbf{Hui Wang\textsuperscript{1,2}},
\\
 \textbf{Zexin Lu\textsuperscript{1,2}},
 \textbf{Yi Zhang\textsuperscript{1,*}}
\\
\\
 \textsuperscript{1}College of Computer Science, Sichuan University\\
 \textsuperscript{2}Key Laboratory of Data Protection and Intelligent Management, Ministry of Education, Sichuan University,\\
 \textsuperscript{3}Centre for Frontier AI Research, Agency for Science, Technology and Research\\
 \textsuperscript{4}School of Cyber Science and Engineering, Sichuan University
\\
 \small{
   \textsuperscript{*}{Corresponding Author:} \href{mailto:Yan_Ming@cfar.a-star.edu.sg}{Yan\_Ming@cfar.a-star.edu.sg} and \href{mailto:yzhang@scu.edu.cn}{yzhang@scu.edu.cn}
 }
}

\usepackage{multirow}
\usepackage{amsmath}
\usepackage{tikz}
\usepackage{subcaption}
\usepackage{subfloat}
\DeclareCaptionFormat{mysubformat}{\fontsize{8}{11}\selectfont#1#2#3}
\usepackage{amssymb}

\newcommand{\mat}[1]{\mathbf{#1}} 

\begin{document}
\maketitle
\begin{abstract}
The surge of hate speech on social media platforms poses a significant challenge, with hate speech detection~(HSD) becoming increasingly critical. Current HSD methods focus on enriching contextual information to enhance detection performance, but they overlook the inherent uncertainty of hate speech. We propose a novel HSD method, named trustworthy hate speech detection method through visual augmentation (TrusV-HSD), which enhances semantic information through integration with diffused visual images and mitigates uncertainty with trustworthy loss. TrusV-HSD learns semantic representations by effectively extracting trustworthy information through multi-modal connections without paired data. Our experiments on public HSD datasets demonstrate the effectiveness of TrusV-HSD, showing remarkable improvements over conventional methods.
\end{abstract}


\section{Introduction}

\begin{figure}
    \centering
    \includegraphics[width=.95\columnwidth]{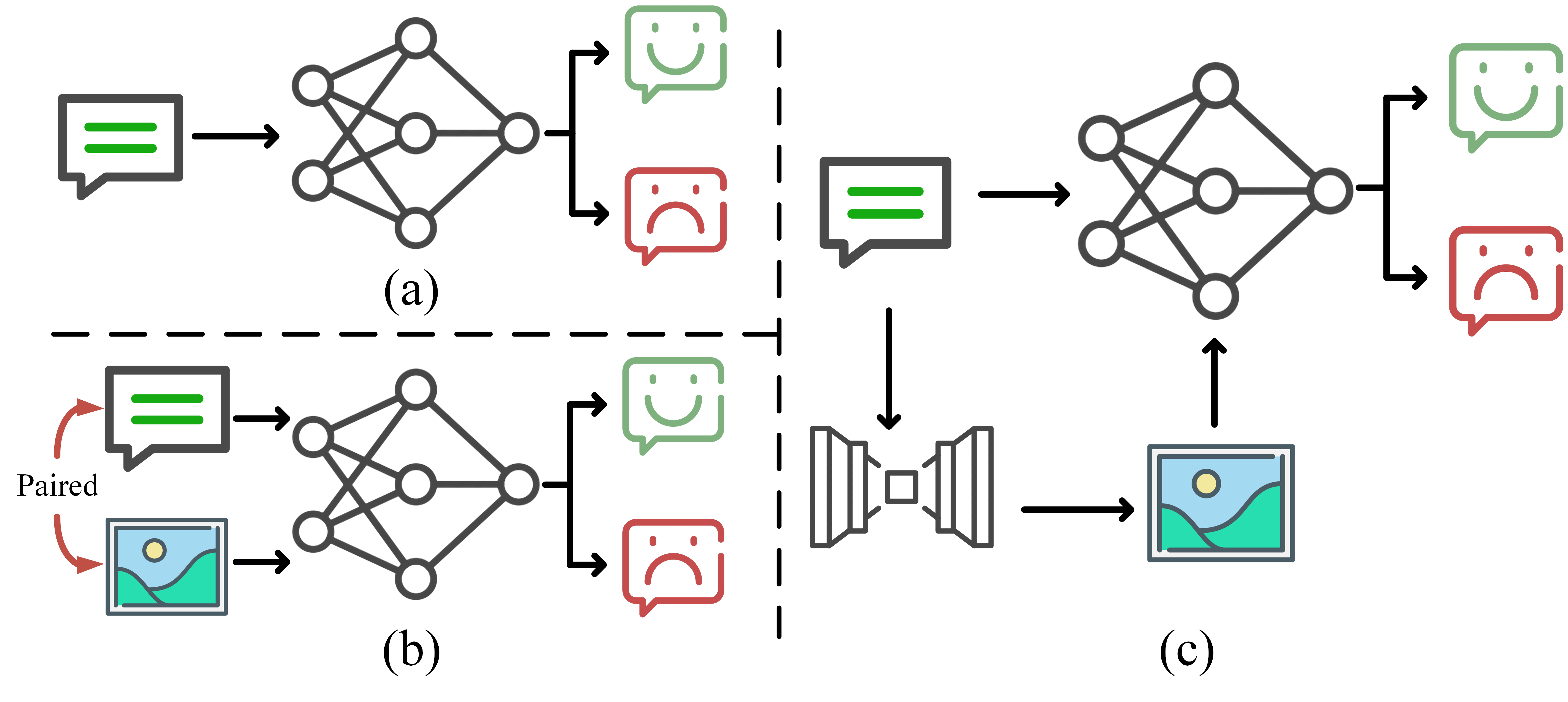}
    \vspace{-10pt}
\caption{Illustration of three HSD paradigms: (a) classic HSD, extracting only text features; (b) multi-modal HSD, requiring paired image and text data; (c) our TrusV-HSD, utilizing only text input alongside generated imaginary for multimodal hate speech detection.}
  \label{fig:Toy}
  \vspace{-20pt}
\end{figure}

The widespread use of social media has both facilitated freedom of individual expression and intensified concerns over the rise of hate speech, emphasizing the urgent need for effective automatic hate speech detection~(HSD) methods~\cite{rawat2024hate, fortuna2022directions}. 
Typically, hate speech mostly appears in short text with very limited contextual information, which increases the detection difficulty~\cite{li2021covid}. Previous HSD methods mostly focused on designing advanced feature representations to improve detection performance: kernel-based methods~\cite{Gamback:Workshop-2017, abro2020automatic}, pre-trained language models~\cite{Sarkar:arxiv-2021, daouadi2023deep}, graph attention networks~\cite{Mishra:arxiv-2019, miao2024gat}. All these methods solely extract features from textual speech (see Figure~\ref{fig:Toy}(a)), which constrained the capability of detection model under limited textual information.

\begin{figure*}
\vspace{-5pt}
\centering
    \includegraphics[width=.9\textwidth]{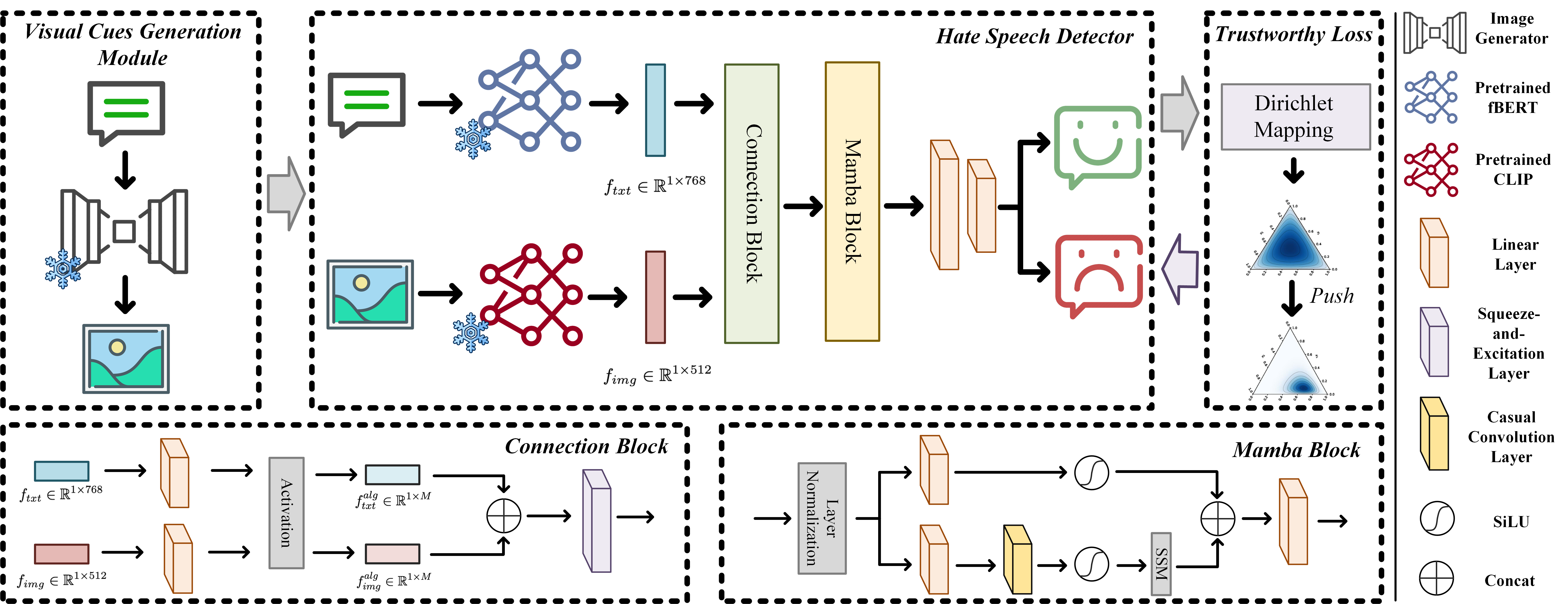}
    \vspace{-5pt}
  \caption{The overview of the proposed method.}
  \label{fig:overview}
  \vspace{-20pt}
\end{figure*}

Another stream of HSD methods~(see Figure~\ref{fig:Toy}(b)) seeks to enrich feature representation from paired textual speech and visual image~\cite{el2024aast, maity2022multitask, bhandari2023crisishatemm}. The polite work~\cite{Boishakhi2021icbd} achieves superior results than sole textual feature-based HSD methods. Similarly, \citet{yang2022multimodal} and \citet{hebert2024multi} enrich domain knowledge with pretrained vision-language models, demonstrating the importance of visual imagery in detecting hate speech. This stream methods heavily rely on paired images and text, which significantly hinders their practical implementation. However, both HSD paradigms overlook the intrinsic length characteristic of hate speech, where shorter speeches often contain more explicit hate semantics, and vice versa. Moreover, the subjective or biased factors from speakers further lead to high uncertainty in predictions. 


To address abovementioned issues, we propose a novel trustworthy hate speech detection method via visual augmentation, dubbed TrusV-HSD. This method comprises three main modules: visual cues generation module, hate speech detector, and trustworthy loss. Inspired by recent research that visual cues can significantly enhance human abilities to retrieve, remember, and understand information~\cite{tan2024audioxtend}, our visual cues generation module initially generates imagery cues. Consquently, the generated imagery cues enhance TrusV-HSD's semantic learning through visual augmentation from textual speech, in a multidimensional way without requiring paired data. Then, a novel multi-modal hate speech detector is designed to model long-range interactions within the joint modalities while maintaining linear computational complexity. Lastly, a trustworthy loss is utilized to provide reliable predictions and ensure robust performance, mitigating the impact of the speaker's subjectivity and bias, as well as ensuring the robust performance of TrusV-HSD across varying speech lengths.

\vspace{-5pt}
\section{Methodology}
Figure~\ref{fig:overview} shows the overview of our method, which contains visual cues generation module, hate speech detector and trustworthy loss. TrusV-HSD will be detailed introduced in the following.

\noindent{\textbf{Visual Cues Generation Module}}
The traditional HSD paradigm can be formulated as $\mathcal{F}(X_{txt})\mapsto Y$, where $X_{txt}$ is the textual speech and $Y$ is the prediction. This paradigm only extract features from textual modality, and its effectivenss is limited by the short speech length. To mitigate this issue, we propose a novel HSD paradigm that generates visual cues to complement the textual data. Notably, differ with previous multimodal HSD methods, our method is free from image-text paired data, ensuring higher flexibility and broader applicability. Our paradigm is formulated as $\mathcal{F}(X_{txt}, \mathcal{G}(X_{txt}))\mapsto Y$, where $\mathcal{G}$ is the generator that produces imagery from the speech. Concretely, a image generator~\cite{Rombach_2022_CVPR} is utilized to achieve the above purpose, which consists of a text encoder, a diffusion model, and an autoencoder. The generated imagery is not merely illustrative but also interprets the speech and presents certain metaphors explicitly. Consequently, TrusV-HSD enriches information through visual augmentation, allowing it to learn concrete image concepts alongside abstract textual knowledge.

\noindent\textbf{Hate Speech Detector}
To efficiently model textual hate speech and its visual agumentation, our TrusV-HSD incorporates a lightweight hate speech detector. This design alleviates the data-intensive reliance and circumvents the needs for fine-tuning heavyweight vison-language backbones, making it ideal for the low-data multimodal HSD. Additionally, our proposed detector aligns long-range multimodal features while maintaining linear computational complexity. 

Specifically, we embed the text and image modalities to get $f_{txt} \in \mathbb{R}^{1\times 768}$ and $f_{img} \in \mathbb{R}^{1\times 512}$ using the pretrained fBERT~\cite{Sarkar:arxiv-2021} and CLIP~\cite{radford2021learning}, respectively. Then, a connection block is designed to connect those two embeddings, represented $f_{txt}^{alg}$ and $f_{img}^{alg} \in \mathbb{R}^{1\times M}$, where $M$ is the alignment dimension. To efficient align multimodal features, we use a Squeeze-and-Excitation Layer~\cite{hu2018squeeze} to encourage attentive learning across multimodal embeddings in TrusV-HSD.


Simillarly, we also introduce the state space sequence model (SSM)~\cite{mamba} in this block, which maps the multimodal feature into a sequence $x_t \mapsto y_t $ through a hidden state $h_t \in \mathbb{R}^N$, where $t$ denotes the timestep and $N$ represents the state size. $\mat{A}$ represents the evolution parameter, $\mat{B}, \mat{C}$ denote the projection parameters. The whole process of SSM is formulated as follows:




\vspace{-10pt}
\begin{equation}
    h_t =\bar{\mat{A}}h_{t-1} + \bar{\mat{B}}x_t, 
     \label{eq:2a}
\end{equation}
\vspace{-15pt}
\begin{equation}
    y_t = \mat{C} x_t, 
    \label{eq:2b}
    \vspace{-3pt}
\end{equation}
where $\bar{\mat{A}}$ and $\bar{\mat{B}}$ are the discrete paramters. Based on the zero-order hold method~\cite{zhu2024vision}, $\bar{\mat{A}} = \mathrm{exp}(\Delta\mathbf{A})$ and $\bar{\mat{B}} = (\Delta\mat{A})^{-1}\mathrm{exp}(\Delta\mat{A}-\mat{I})\cdot \Delta\mat{B}$, and $\Delta$ denotes the timescale parameter. 

Lastly, our hate speech detector delivers the output through a global convolution operation, expressed as $g=a* \overline{\mat{K}}$. $\overline{\mat{K}}=\left(\mat{C} \overline{\mat{B}}, \mat{C} \overline{\mat{A B}}, \ldots, \mat{C} \overline{\mat{A}}^{T-1} \overline{\mat{B}}\right)$ is a casual convolutional kernel, and $T$ denotes the input length.

\noindent \textbf{Trustworthy Loss}
Due to subjective or biased factors of speakers and domain independence among individual speeches, it is critical need for network reliability against uncertainties to ensure trustworthy detections.
Inspired by~\citealp{DST} and~\citealp{qin2022deep}, we introduce a trustworthy loss to emphasize highly reliable predictions while weakening the impact of unreliable ones. More discussions can be found in Appendix~\ref{sec:sec:app_trus_loss}.


Given an input $x_i=\{x_{txt},x_{img}\}$, and
the evidence vector $e_i$ is obtained through the network $\mathcal{M}$, formulated as $e_i = \mathcal{M}(x_i)$, where $e_i \ge 0$. 
Following the principle of Subjective Logic (SL)~\cite{SL} and the uncertainty quantification can be written as $u_i + \sum^{K}_{k=1}b_i^k = 1$, where $K$ is the number of classes, $b_i$ is the belief mass and $b_i^k=\frac{e_i^k}{S_i}$. $u_i$ is the uncertainty quantification and $u_i = K/S_i$, where $S_i=\sum^{K}_{k=1} (e_i^k + 1)$.
The belief mass assignment corresponds to a Dirichlet distribution with parameters $\alpha_i$, where $\alpha_i=e_i + 1$. Then, we reformulatethe cross-entropy loss as:
\begin{equation}
    \mathcal{L}_{digamma} =\sum_{k=1}^{K} y_{i}^{k}\left(\psi\left({S}_{i}\right)-\psi\left({\alpha}_{i}^{k}\right)\right),
\tag{3}
\end{equation}
where $\psi(\cdot)$ is the \textit{digamma} function. Moreover, Kullback-Leibler (KL) divergence is incorporated to penalize the divergence from negative samples, which can be defined as:
\begin{equation}
    \mathcal{L}_{K L}
    =K L\left[
    D\left({p}_{i}\mid \tilde{{\alpha}}_{i}\right)
    \|
    D\left({p}_{i} \mid {1} \right)\right] ,
     \tag{4}
\end{equation}
where ${p}_{i}$ is the classification probability and ${p}_{i}=\frac{\alpha_i}{S_i}$. $D(p_i|{1})$ is the uniform Dirichlet distribution, and $\tilde{\alpha}_{i}=y_i+(1-y_i) \odot \alpha_i$. $\mathcal{L}_{KL}$ encourages the model to give negative samples higher uncertainty. Finally, the overall loss can be formulated as:

\begin{equation}
    \mathcal{L}_{trust} = \mathcal{L}_{digamma} + \lambda \mathcal{L}_{KL}, \tag{5}
\end{equation}
where $\lambda$ is a annealing coefficient, which is set to $\lambda(t)=min(1, \frac{c}{0.5 C})$. $c$ is current epoch number, and $C$ is the total training epoch number.

\begin{table*}[htbp]\centering
\vspace{-10pt}
\resizebox{.95\linewidth}{!}{
\begin{tabular}{cccccccc}
\hline
\multirow{2}{*}{\textbf{Models}}  &\multicolumn{3}{c}{\textbf{SE}} & &\multicolumn{3}{c}{\textbf{FNUC}} \\ \cline{2-4} \cline{6-8}
&\textbf{F1}  &\textbf{Precision}  &\textbf{Recall} &  &\textbf{F1}  &\textbf{Precision} &\textbf{Recall} \\ 
\hline
RNNLM~\cite{mehdad2016characters} & 70.62 $\pm$ 4.71 & 71.58 $\pm$ 5.21 & 70.79 $\pm$ 4.42 & & 60.64 $\pm$ 7.41 & 66.86 $\pm$ 7.73 & 60.29 $\pm$ 6.31\\
SVM~\cite{Davidson:AAAI_2017}	  &69.22 $\pm$ 5.22 &69.74 $\pm$ 4.89 &69.42 $\pm$ 4.68 & &57.26 $\pm$ 4.44	&58.03 $\pm$ 5.20	&57.06 $\pm$ 4.24\\
HybridCNN~\cite{Park:arxiv-2017} &70.11 $\pm$ 3.92 &71.40 $\pm$ 4.32 &70.23 $\pm$ 4.85 & &62.11 $\pm$ 5.19 &62.43 $\pm$ 5.94 &62.02 $\pm$ 4.99\\
 \hdashline
BERT~\cite{Devlin:arxiv-2018bert}	&72.11 $\pm$ 3.37	&71.23 $\pm$ 4.01	&73.28 $\pm$ 4.09 &	&62.86 $\pm$ 2.02	&66.81 $\pm$ 3.62	&62.22 $\pm$ 1.91\\
HateBERT~\cite{Caselli:arxiv-2020} &72.15 $\pm$ 4.86 &68.26 $\pm$ 6.78 &74.41 $\pm$ 4.53 & &62.27 $\pm$ 5.16 &64.29 $\pm$ 4.93 &62.05 $\pm$ 4.97 \\
fBERT~\cite{Sarkar:arxiv-2021}	&71.37 $\pm$ 5.97	&75.84 $\pm$ 5.74	&71.68 $\pm$ 5.49 &	&62.41 $\pm$ 2.83	&64.63 $\pm$ 3.99	&61.75 $\pm$ 2.52\\ \hdashline
RGCN~\cite{schlichtkrull2018modeling} & 73.33 $\pm$ 3.79 & 73.71 $\pm$ 3.61 & 73.38 $\pm$ 3.67 & & 68.92 $\pm$ 5.75 & 68.86 $\pm$ 5.78 & 70.12 $\pm$ 6.08\\
GCN~\cite{Mishra:arxiv-2019}	&66.13 $\pm$ 4.11	&66.57 $\pm$ 4.32	&69.26 $\pm$ 8.63&	&55.50 $\pm$ 3.90	&55.82 $\pm$ 4.23	&57.06 $\pm$ 4.62\\
RSGNN~\cite{Song:IPM_2022}	&74.04 $\pm$ 4.01	&74.29 $\pm$ 3.88	&74.14 $\pm$ 3.85 &	&69.54 $\pm$ 5.36	&\textbf{70.15 $\pm$ 5.48}	&69.97 $\pm$ 5.59\\
GAT~\cite{miao2024gat}	&-	&-	&- &	&64.45 $\pm$ 3.54	&63.66 $\pm$ 3.45	&67.26 $\pm$ 3.78\\\hdashline
AAST-NLP~\cite{el2024aast} & 76.34 $\pm$ 4.49 & 76.24 $\pm$ 3.87 & 77.74 $\pm$ 4.56 & &64.42 $\pm$ 4.19 & 64.06 $\pm$ 4.05 & 65.40 $\pm$ 4.59\\
 \hline
TrusV-HSD	&\textbf{78.92 $\pm$ 4.26}	&\textbf{78.80 $\pm$ 4.05}	&\textbf{79.44 $\pm$ 4.51} &	&\textbf{70.27 $\pm$ 3.87} 	& 69.37 $\pm$ 3.79 &\textbf{72.77 $\pm$ 4.61}\\
\hline
\end{tabular}
}
\vspace{-5pt}
\caption{The average results of different approaches along with the standard deviation (\%). }
\vspace{-15pt}
\label{tab:results}
\end{table*}

\section{Experiments}

\noindent \textbf{Experimental Environment}
TrusV-HSD is implemented on PyTorch, optimized by Adam~\cite{kingma2014adam} with a learning rate of 0.001. The training epoch is set to 800. The experimental environment includes an AMD Ryzen 7 5800X CPU and four NVIDIA GTX 3080 Ti GPUs.

\noindent \textbf{Datasets and Metrics}
Two public explicit HSD datasets are used to validate the proposed method, including SemEval2019 task-5 (SE)~\cite{Basile:SE-2019} and Fox News User Comments~(FNUC)~\cite{Gao:FUNC-2017} datasets. We also validate our method in the implicit HSD task using the public Implicit Hate Speech (IHC)~\cite{elsherief-etal-2021-latent} dataset. SE, FUNC and IHC contain 12,000, 1,528 and 22,056 samples, respectively. We use 10-fold cross-validation to validate the methods. 
Due to serious sample imbalance, we use F1, Precision, and Recall as evaluation metrics. The average results and standard deviations are presented for each experiment.
\noindent \textbf{Explicit Hate Speech Detection}
The results of different methods on SE and FNUC are presented in Table~\ref{tab:results}. Benifiting from integration of visual cues and trustworthy loss, our TrusV-HSD achieves the best F1 scores in the comparisons with other methods. Our TrusV-HSD also achieves optimal or near-optimal precision and recall, with notable improvements in SE dataset.
In FNUC dataset, TrusV-HSD exhibits superior F1 performance, along with a significant increase in recall (+2.80\%), although the precision is slightly lower (-0.78\%) compared to RSGNN. This variation is likely due to the small number of samples in FNUC. 
Besides, our method outperforms AADT-NLP, a multi-modal HSD method, demonstrating the effectiveness of our proposed detector and the trustworthy loss.

\noindent \textbf{Implicit Hate Speech Detection}
The above experiments validate our method's effectiveness in detecting explicit hate speech. Implicit hate speech, masked by subtle tricks, is more challenging to detect. To assess our method comprehensively, we also evaluate its performance in implicit hate speech detection task, focusing on reasoning ability and robustness (see Table~\ref{tab:ihsd_results}). Due to severe data imbalance in IHC dataset, BERT tends to predict samples as positive, resulting in high precision but low recall. Despite this, our significant F1 score improvement demonstrates TrusV-HSD's superiority in large-scale datasets by effectively learning connections between modalities.





\begin{table}[!t]\centering
\small
\resizebox{\linewidth}{!}{
\begin{tabular}{cccc}
\hline
\multirow{2}{*}{\textbf{Models}}  &\multicolumn{3}{c}{\textbf{IHC}}\\ \cline{2-4}   &\textbf{F1}  &\textbf{Precision} &\textbf{Recall} \\ 
\hline
BERT	  &67.83 $\pm$ 7.61 & 71.05 $\pm$ 6.44 & 68.49 $\pm$ 7.73 \\
fBERT	  &65.32 $\pm$ 5.42 &65.52 $\pm$ 5.34 & 65.30 $\pm$ 5.48 \\
GCN	  & 68.40 $\pm$ 3.93 & 68.14 $\pm$ 3.77 &69.09 $\pm$ 4.23 \\
RSGNN	  &69.22 $\pm$ 5.22 &69.74 $\pm$ 4.89 &69.42 $\pm$ 4.68 \\
AAST-NLP	  & 67.48 $\pm$ 3.10 &67.39 $\pm$ 3.16 &67.79 $\pm$ 2.99 \\\hline
TrusV-HSD	  &\textbf{71.19$\pm$ 4.30} & \textbf{70.88 $\pm$ 4.32} & \textbf{71.78 $\pm$ 4.23} \\ 
\hline
\end{tabular}
}
\caption{The average results of different approaches along with the standard deviation (\%) in IHC.}
\vspace{-20pt}
\label{tab:ihsd_results}
\end{table}




\noindent \textbf{Ablation Study}
The ablation study is tested in SE dataset, and the results are shown in Table~\ref{tab:abla}, where "Detector$\dagger$" and "Detector$\ddagger$" denote our detector without and with the connection block, respectively\footnote{Detailed seetings can be found in Appendix~\ref{sec:app_exp_set}}. We treat fBERT as the baseline, an significant improvement can be observed when imagery cues are added ("+Imagery"). However, simply replacing the classification head ("+Detector$\dagger$") cannot improve performance, as the relationships between modalities are not well connected. Adding the connection block ("+Detector$\ddagger$") improves performance, highlighting the importance of connecting modalities. Morever, introducing the trustworthy loss enhances performance through addressing the domain independence problem. Besides, please check generation experiments in Appendix~\ref{sec:app_gen_exp}.

\begin{figure}[!h]
\vspace{-10pt}
    \centering
    \includegraphics[width=.8\columnwidth]{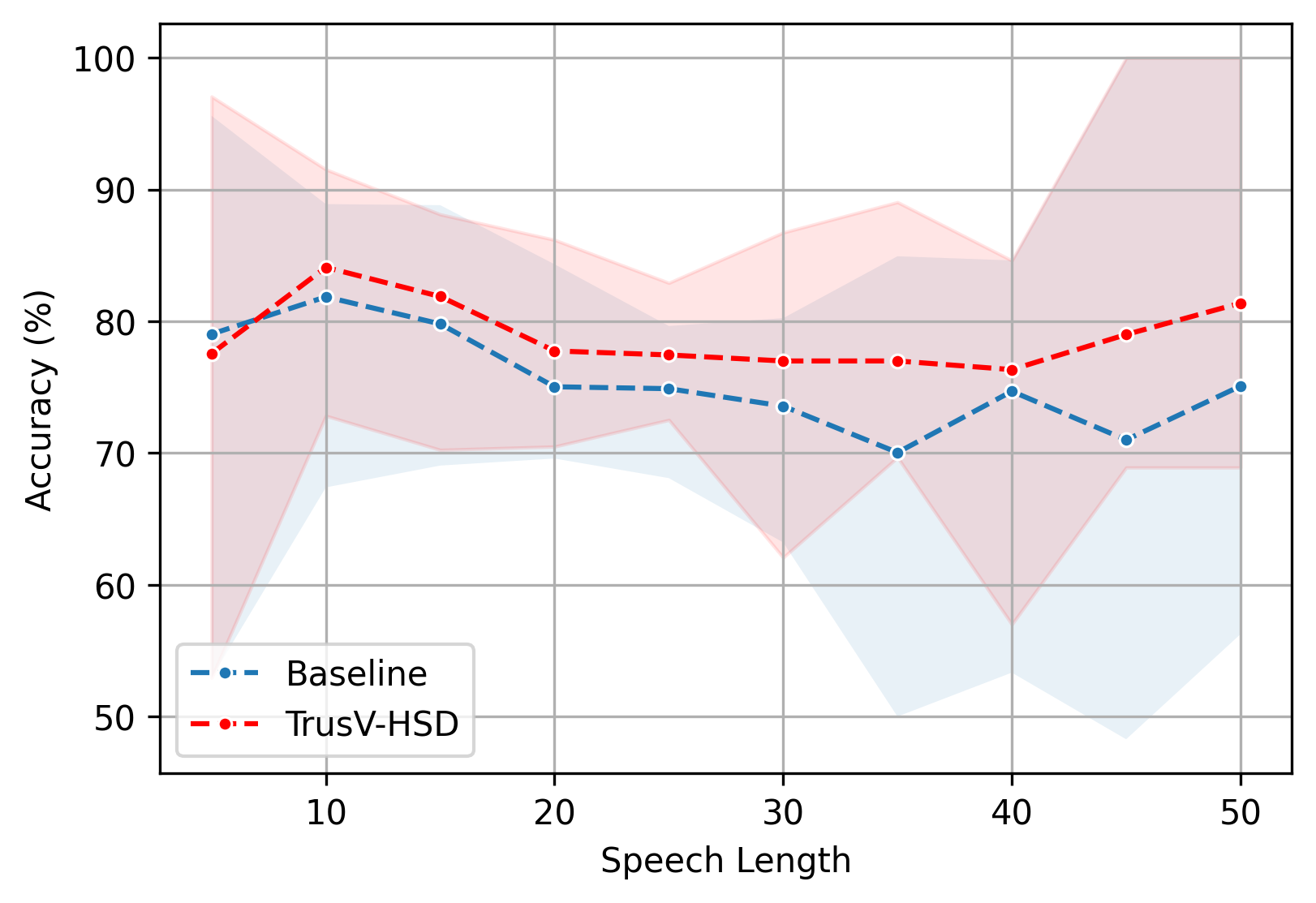}
    \vspace{-5pt}
    \caption{
The average performance and confidence intervals across different speech length bins.}
    \vspace{-10pt}
    \label{fig:comparison}
\end{figure}

Figure~\ref{fig:comparison} illustrates the average performance and confidence intervals of TrusV-HSD and baseline (TrusV-HSD without trustworthy loss) across various speech lengths in SE. It can be observed that the performance of the baseline declines with increasing speech length. This is because longer texts contain more of the speaker's subjectivity, leading to significant domain independence problems. TrusV-HSD consistently outperforms the baseline in terms of accuracy across all lengths. This demonstrates that the trustworthy loss helps TrusV-HSD achieve promising robustness and reliability, effectively handling diverse speech lengths and alleviating the domain independence problem.


\footnotetext[1]{The hate data is sourced from public datasets or generated by a pretrained image generator. These contents do not reflect our views, and we sincerely apologize for any offense caused.}

\begin{table}[!t]\centering
\small
\resizebox{\linewidth}{!}{
\begin{tabular}{cccc}
\hline
\multirow{2}{*}{\textbf{Models}}  &\multicolumn{3}{c}{\textbf{SE}}\\ \cline{2-4}   &\textbf{F1}  &\textbf{Precision} &\textbf{Recall} \\ 
\hline
baseline	&71.37 $\pm$ 5.97	&75.84 $\pm$ 5.74	&71.68 $\pm$ 5.49 \\
+ Imagery	  &76.34 $\pm$ 4.49 & 76.24 $\pm$ 3.87 & 77.74 $\pm$ 4.56 \\
+ Detector$\dagger$	  &75.28 $\pm$ 3.68 & 75.21 $\pm$ 3.47 & 75.63 $\pm$ 3.84 \\
+ Detector$\ddagger$	  &77.64 $\pm$ 4.03 & 77.34 $\pm$ 3.87 & 78.62 $\pm$ 4.28 \\
+ Trustworthy & 78.92 $\pm$ 4.26	& 78.80 $\pm$ 4.05	& 79.44 $\pm$ 4.51 \\
\hline
\end{tabular}
}
\vspace{-5pt}
\caption{The average results of ablation studies along with the standard deviation (\%) in SE.}
\vspace{-5pt}
\label{tab:abla}
\end{table}



\begin{figure}[!t]
    \includegraphics[width=\columnwidth]{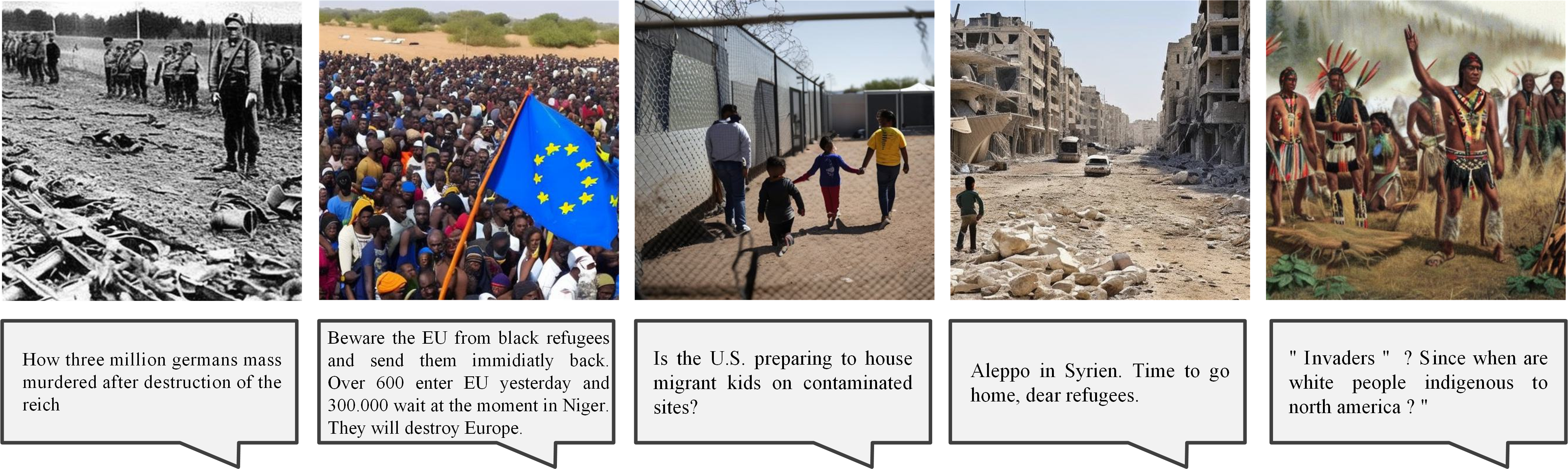}
    \vspace{-15pt}
  \caption{The visualized cues\protect\footnotemark[1].}
  \label{fig:visual}
  \vspace{-20pt}
\end{figure}

\noindent \textbf{Visualization Examples}
Several visualized samples are shown in Figure~\ref{fig:visual}, it can be noticed that our generation module enrich semantic information by generating concrete imagery from textual speech. Besides, the generated images could present certain metaphors explicitly, as shown in the final sample of Figure~\ref{fig:visual}. Thereby, the imagery cues help the model learn concrete image concepts alongside abstract textual knowledge.

\vspace{-5pt}
\section{Conclusion}

In this paper, we propose a novel trustworthy hate speech detection method via visual augmentation. Concretely, our method can generate imagery cues to enrich information from brief speech, and the designed lightweight detector can effectively alleviate the data-intensive reliance issue. Finally, the trustworthy loss enhances robustness and ensures reliable predictions across diverse speech lengths, alleviating the domain independence problem. In the future work, one possible direction is modeling user-specific features to detect hate speech.

\section{Limitations}

While our proposed TrusV-HSD shows promising results in both explicit and implicit hate speech detection tasks, it is not without limitations. 

\noindent \textbf{Computational Overhead:} The process of generating images from the speech adds computational complexity and time. 

\noindent \textbf{Potential Risk:} Like many AI systems, TrusV-HSD could be susceptible to adversarial attacks designed to manipulate its outcomes. 

\noindent \textbf{Language Limitation:} Our method has only been tested on English speech. Its effectiveness and accuracy in detecting hate speech in other languages remain unverified, which may limit its applicability in multilingual settings.

\noindent \textbf{Bias and Fairness}: While the trustworthy loss aims to reduce subjectivity and bias, it may not completely eliminate them. One possible solution is to provide the speaker's prior, but related dataset is not collected yet.

\section{Ethical Considerations}

In this study, we emphasize that all datasets used are publicly available and have been sourced from established repositories. The content within these datasets, including any hate speech examples, does not reflect the views or opinions of the authors. Our research aims to advance the field of hate speech detection for the benefit of the community, and the use of such data is solely for academic and research purposes. Any offensive or harmful content present in the datasets is used only to develop and test our detection models, and we do not endorse or support any of the sentiments expressed in the data.

The image generator was primarily trained on subsets of LAION-2B~\cite{schuhmann2022laion}, which consists mainly of images with English descriptions. Consequently, texts and images from non-English communities and cultures are underrepresented, often defaulting to white and western perspectives. This limitation affects the model's ability to generate content with non-English prompts, producing significantly poorer results. Stable Diffusion v2 amplifies these biases to a degree that viewer discretion is advised, regardless of the input or its intent.

\bibliography{acl_latex-short}

\appendix

\section{Appendix}
\label{sec:appendix}
In the appendix, we first present the motivation behind this paper, followed by a detailed discussion of the trustworthy loss and its effectiveness in this task. Next, we describe the experimental setup. Finally, we conduct a generation experiment to demonstrate the flexibility of the proposed method.

\subsection{Motivation}
Our main motivation is to introduce a novel learning paradigm for the hate speech detection task by leveraging feature learning from both text and vision modalities simultaneously, rather than proposing a specific framework. This approach differs from previous paradigms that rely solely on the text modality. Additionally, our learning paradigm does not require paired text-image data, ensuring higher flexibility and broader applicability.

We also want to highlight that all the pre-trained feature extraction components in the proposed TrusV-HSD are pluggable and replaceable. In other words, the performance of our method could be further improved with more advanced text or visual pre-trained backbones in the future.

\begin{table*}[!h]\centering
\resizebox{\linewidth}{!}{
\begin{tabular}{cccccccc}
\hline
\multirow{2}{*}{\textbf{Models}}  &\multicolumn{3}{c}{\textbf{SE}} & &\multicolumn{3}{c}{\textbf{FNUC}} \\ \cline{2-4} \cline{6-8}
&\textbf{F1}  &\textbf{Precision}  &\textbf{Recall} &  &\textbf{F1}  &\textbf{Precision} &\textbf{Recall} \\ 
\hline
BERT~\cite{Devlin:arxiv-2018bert}	&72.11 $\pm$ 3.37	& 77.36 $\pm$4.92	& 77.67 $\pm$4.21 &	& 69.39 $\pm$3.37	& 69.45 $\pm$3.27	& 69.82 $\pm$3.99\\
TrusV-HSD* &77.93 $\pm$ 4.12 &68.26 $\pm$ 6.78 &74.41 $\pm$ 4.53 & &62.27 $\pm$ 5.16 &64.29 $\pm$ 4.93 &62.05 $\pm$ 4.97 \\
\hdashline
fBERT~\cite{Sarkar:arxiv-2021}	&71.37 $\pm$ 5.97	&75.84 $\pm$ 5.74	&71.68 $\pm$ 5.49 &	&62.41 $\pm$ 2.83	&64.63 $\pm$ 3.99	&61.75 $\pm$ 2.52\\ 
TrusV-HSD\#	&\textbf{78.92 $\pm$ 4.26}	&\textbf{78.80 $\pm$ 4.05}	&\textbf{79.44 $\pm$ 4.51} &	&\textbf{70.27 $\pm$ 3.87} 	& 69.37 $\pm$ 3.79 &\textbf{72.77 $\pm$ 4.61}\\
\hline
\end{tabular}
}
\caption{The results of TrusV-HSD with different textual backbones (\%). }
\label{tab:results}
\end{table*}

\subsection{Trustworthy Loss}
\label{sec:sec:app_trus_loss}
Different speech lengths can lead to various distributions, causing the model to struggle with this issue. A similar problem also occurs with other parameters, such as for people of different regions or genders. If the test distribution differs from the training distribution (e.g., different races, genders, religions), it results in a typical out-of-distribution (OOD) problem. Numerous studies~\cite{DST,chen2024evidence} have demonstrated that the trustworthy loss can effectively handle this problem by replacing this parameter set with the parameters of a Dirichlet density. The trustworthy loss transforms the model's predictions into a distribution over possible softmax outputs, rather than a point estimate of a softmax output. In this way, our model could measure that if the test data is in or out of the distribution.

Specifically, the traditional classification loss classifies the test data into known classes based on the prior knowledge of the training data. However, if the test and training data follow different distributions, this process may lead to misclassification. The trustworthy loss can effectively alleviate this problem by quantifying the uncertainty~\cite{li2022trustworthy}. This approach allows our model to reduce the high reliance on prior knowledge and provide a trustworthy prediction.

Our experiment in our manuscript, Figure 3 in the main text, supports the above statement. 86.44\% of the speech in SE is less than 35 words, leading to the model overfitting to the prior knowledge of the high-frequency data distribution. Consequently, standard classification loss (cross-entropy loss)-based methods exhibit a significant performance gap between short and long speeches. Our trustworthy loss effectively alleviates this problem, ensuring the model maintains performance across different data distributions (short and long speeches).

\subsection{Experiment Setting}
\label{sec:app_exp_set}
In our paper, 10-fold cross-validation is used to evaluate the robustness and performance of our model comprehensively. This way reduces the variance associated with random partitioning of the data and provides a more thorough assessment of our model's performance across different subsets, leading to more reliable and stable results. Meanwhile, 10-fold cross-validation is a classical validation protocol in hate speech detection~\cite{Song:IPM_2022}, so we chose to validate both our method and the compared methods using this approach.

In the ablation study, our multi-modal baseline has linear layers (without Connection block and Mamba block) after the feature concatenation operation. "Detector$\dagger$" refers to replacing one linear layer with a Mamba Block but without the connection block. "Detector$\ddagger$" refers to use both Mamba and connection blocks. "Detector$\dagger$" and "Detector$\ddagger$" are optimized with cross-entropy loss, not the trustworthy loss.

\subsection{Generation Experiment}
\label{sec:app_gen_exp}
To evaluate the generalization performance and illustrate that our proposed method is a pretrained backbone-free method, we evaluated our methods with different backbones: TrusV-HSD\# (with BERT) and TrusV-HSD* (with fBERT). The results, as shown in Table 1, clearly indicate that our proposed framework significantly improves detection performance on two public datasets, achieving a 5\% F-1 score improvement with BERT and a 7\% F-1 score improvement with fBERT in the SE dataset.

Mamba can effectively model long-range interactions within the joint modalities while maintaining linear computational complexity~\cite{mamba}. For the multi-modal feature extractor, the long-range relationship between the features of the two modalities is crucial. Therefore, in this paper, we utilize Mamba to extract discriminative multi-modal features. Related ablation experiments with and without Mamba can be found in our manuscript (Table 3). Simply replacing the classification head (linear layer) with the Mamba block did not improve performance, likely because the latent relationships between the two modalities were not well connected. In this finding, we carefully designed the connection module to align the features of the two modalities.


\end{document}